IAC–24–B.6.2.IP

# Machine learning-driven Anomaly Detection and Forecasting for Euclid Space Telescope Operations

**Pablo Gómez[a*], Roland D. Vavrek[a], Guillermo Buenadicha[a], John Hoar[a], Sandor Kruk[a], Jan Reerink[a]**

[1] *European Space Agency (ESA), European Space Astronomy Centre (ESAC), Camino Bajo del Castillo s/n, 28692 Villanueva de la Cañada, Madrid, Spain,* pablo.gomez@esa.int

[*] *Corresponding author*

## Abstract

State-of-the-art space science missions depend to an increasing degree on automation due to the complexity of the spacecraft and due to operational costs induced by human oversight and intervention. Additionally, the large amounts of data generated by the spacecraft, both in terms of scientific output but also in terms of telemetry data, are on a scale that makes human inspection and interpretation of the data challenging. Machine learning-based methods have the potential to play a significant role in this area. The Euclid space telescope, which entered its survey phase in February 2024, is a prime example of this. The groundbreaking scientific output that Euclid will provide relies on the correct interpretation and adaptation to the spacecraft's housekeeping telemetry data and data derived from science processing. Spacecraft housekeeping data encompasses thousands of parameters that are monitored as a time series. These parameters may or may not have direct impact on the quality of the generated science output and have complex interactions depending on physical relationships (e.g. nearby positions of temperature sensors). To ensure optimal science operations a detailed study of anomalies recorded and identifying anomalous parameter states that are hidden in the data is essential. Secondly, understanding the relationship between known anomalies and the monitored physical quantities is critical but non-trivial as related parameters may display anomalous behaviour at different points in time and at different strength. In this paper, we address these challenges by analysing temperature anomalies in Euclid's telemetry data collected during its survey phase from February 2024 to August 2024. In particular, we study eleven temperature parameters and 35 covariates, for which we are interested in their interactions with the temperatures. We utilise an approach that combines a predictive XGBoost model that predicts the temperatures based on previous values at various time lags. We then identify anomalies in relation to the prediction to detect deviations from expected temperature behaviour and train a second XGBoost model to predict anomalies based on the covariates to learn the relationship of covariates and anomalies. For each parameter, we identify the top three anomalies and investigate their interactions with the covariate parameters using SHAP (Shapley Additive Explanations). This allows us to automatically and rapidly assess complex parameter relationships and gain insights into the underlying causes of these anomalies. Our method demonstrates how machine learning can enhance the monitoring and interpretation of large-scale spacecraft telemetry, offering a scalable solution that has the potential to be applied to other missions facing similar data challenges.

**Acronyms/Abbreviations**

| | |
|---|---|
| AI | Artificial Intelligence |
| ESA | European Space Agency |
| SHAP | Shapley Additive Explanations |
| VIS | Euclid's Visible Light Instrument |
| XGBoost | eXtreme Gradient Boosting |

## 1. Introduction

Modern space missions, such as those conducted by the European Space Agency (ESA), are becoming increasingly complex, producing vast amounts of data both from scientific instruments and other spacecraft systems. The need for automation is critical, as manually monitoring and analysing the data is costly and only possible to a limited degree given the scale of the data. This is also highlighted by ESA's Artificial Intelligence for Automation (A2I) Roadmap [1], which has emphasised the importance of artificial intelligence (AI) in automating mission operations, where AI-based methods are seen as a key enabler of efficient, cost-effective space exploration.

One key topic is the detection of anomalies in telemetry data which plays a crucial role in ensuring the operational stability of spacecraft. Telemetry data encompasses numerous parameters, and identifying significant deviations in these parameters is critical for early warning systems, preventative maintenance and maximising science return from instruments. This is highlighted, e.g., by the recent work by Kotowski *et al.* [2], who provide a high-quality public dataset with anomalies and telemetry from multiple spacecraft. A challenge, however, lies not only in detecting anomalies but in understanding the relationships between parameters that may drive these deviations. In-terpretable machine learning, specifically techniques like explainable AI, can offer insights into which factors contribute to anomalies and provide actionable information for mission operators.

The Euclid mission is a prime example of a complex space science mission that has seen developments in this area through its health monitoring system [3]. Launched by ESA to explore dark energy and dark matter, Euclid entered its survey phase in February 2024, and its scientific success heavily depends on the optimal functioning of its instruments, which must be monitored in near real-time [4]. Especially temperature relationships have proven central for Euclid, thermo-elastic variations cause tiny variations in image quality biasing galaxy shape measurements. Thus, given the importance of thermal control and temperature stability in space operations, anomalies in temperature readings can have a direct impact on the mission's scientific output. Temperature anomalies can disrupt the quality of scientific data and lead to operational risks.

In this paper, we present an automated approach for detecting and analysing temperature anomalies in Euclid's telemetry data. Our method leverages machine learning techniques, including XGBoost for anomaly detection and Shapley Additive Explanations (SHAP) for interpreting the relationships between various telemetry parameters [5, 6]. We demonstrate how predictive models based solely on covariate data can effectively detect significant deviations in temperature parameters over time. By applying SHAP, we offer a transparent interpretation of how different covariates drive model predictions, providing a more detailed understanding of the interactions between parameters and providing potential insight on causal relationships. The key contributions of this work include:

- Demonstrating the effectiveness of predictive models in detecting temperature anomalies in Euclid data using only covariate data.
- Providing a robust methodology for automated anomaly detection and interpretation using SHAP values.
- Identifying key anomalies in the monitored time frame and studying the underlying factors.

Our approach represents a scalable and adaptable method for anomaly detection in spacecraft telemetry, which can be applied to other missions facing similar challenges in handling large-scale telemetry data.

## 2. Materials and Methods

### *2.1 Dataset*

In this study, we analyse telemetry data collected from the Euclid spacecraft during its survey phase from February 15, 2024, to August 14, 2024. The dataset consists of 11 temperature parameters and 35 covariate parameters, sampled every 10 seconds, resulting in a total of 1,572,480 sample points per parameter. The 11 temperature sensors are all on the payload module of the spacecraft as illustrated in Figure 1. The covariate parameters are described in Table 1.

Table 1. Overview of analysed covariate parameter types used

| Name / Type | Description |
|---|---|
| GACS9041 | Solar aspect angle that is the angle between the spacecraft +ZS/C axis (telescope pointing direction) and the direction to the centre of the solar disk. |
| GFDS9161 | Alpha angle, angle between the Sun vector projected onto the XS/C-YS/C plane and the +XS/C axis. |
| NIST06xx | NISP instrument internal temperatures |
| NIST03xx | NISP instrument wheel currents |
| PPPDxxxx | Power values for different instrument units |
| SSSTxxxx | Thermal sensors on the ring connecting instrument cavity to warm electronics and spacecraft bottom |
| SSSDxxxx | Thermal sensors on various positions the PLM Baseplate (SSSD0172 is on the VIS radiator) |
| VMGT3034 | VIS instrument shutter motor current |

To reduce the volume of data and since temperature parameters are typically not highly erratic, we resample the data to a 10-minute sampling interval, i.e. 26,064 data points per parameter.

Plots of the 11 temperature parameters and the 35 covariate parameters are shown in Figures 2 and 3. These visualisations provide an overview of the behaviour of the telemetry data over the time span of the study. Correlation heatmap of the covariate parameters with each other and observed temperatures are also provided in Figures 4, which helps visualise the relationships between the covariates. While we can see some strong correlations, especially with NIST06xx parameters, the other parameters show no strong correlations.

### *2.2 Anomaly Detection Methodology*

The primary goal of this study is to detect and analyse temperature anomalies in the telemetry data to aid in bet-ter understanding the heat flow within the spacecraft. To achieve this, we utilise a two-model approach combining predictive modelling and anomaly scoring based on the *darts* Python module [7] and performing anomaly prediction with the second model. For analysis we rely SHAP values to interpret the second model's predictions.

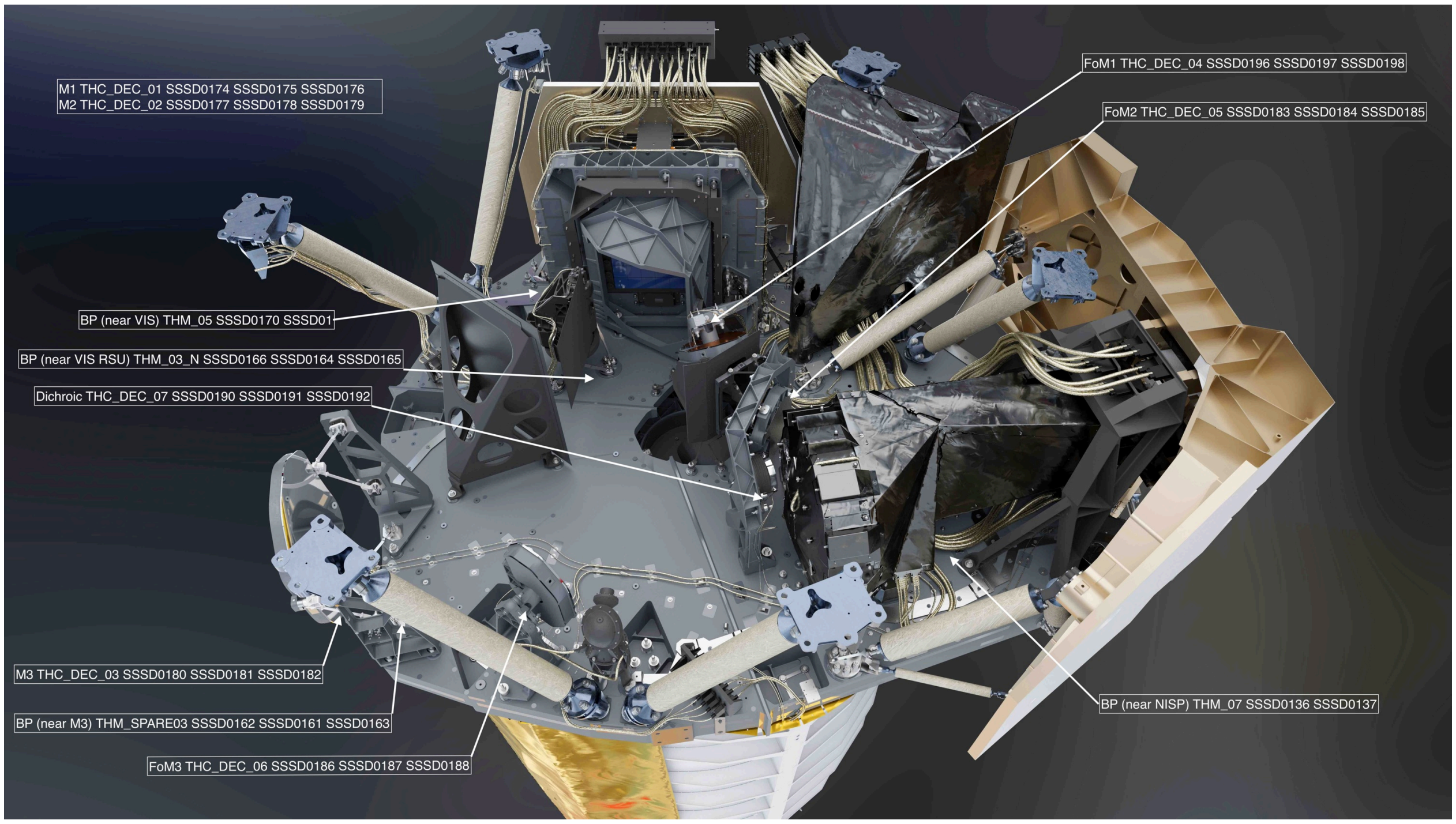

Fig. 1. Instrument cavity of Euclid with positions of some of the temperature sensors being indicated on the lower side of the baseplate. (Credit: ESA)

#### *2.2.1 Model 1: Temperature Prediction and Anomaly Scoring Based on Previous Temperature Values*

The first model is designed to predict future temperature values based solely on the previous values of the same temperature parameter. This model is constructed using a gradient-boosted decision tree algorithm (XGBoost). XGBoost is selected due to its ability to handle large datasets efficiently and its robustness in modelling time series data. The model takes the temperature values at two time lags —- 15 minutes and 1 hour prior to the prediction point —- as input features. The model is trained independently for each of the 11 temperature parameters in the dataset. By focusing only on previous values of the same temperature parameter, the model is forced to learn the inherent temporal patterns and short-term dependencies in the temperature data. This model is trained on the first 66% of the data, i.e. on 17,202 training data points.

Once trained, the model generates temperature predictions for each parameter over the study period. These predictions are compared to the actual observed temperatures, and the differences (residuals) between the predicted and actual values are treated as potential anomalies. Any significant deviation between the predicted and actual temperature values is indicative of a possible anomaly. To score the anomalies, we apply a k-means clustering method on the residuals. A window size of 64 samples is used to cluster residuals and identify significant deviations from normal behaviour. Any residuals that fall outside the expected clusters are flagged as anomalies. This clustering method allows us to identify periods where the temperature deviated from its expected value in a systematic way. The anomalies detected by this model serve as input for the second model, which investigates the relationships between the anomalies and the other covariates.

#### *2.2.2 Model 2: Residual and Anomaly Prediction Based on Covariates*

The second model is designed to allow us inferring the causes of the temperature anomalies by predicting the residuals (i.e., the differences between the predicted and actual temperature values from Model 1) based solely on the covariate parameters. This model also uses XGBoost. Three time lags for the covariate parameters are used as inputs to the model, with three data points given up to each lag at 30 minutes, 4 hours and 24 hours. This enables the model to capture both short-term and long-term dependencies between the covariates and the anomalies. This model is fit on the remaining 34%, i.e. 8,862 test data points for each parameter. By excluding the temperature parameters as input, this model is tasked with learning the relationships between the covariates and the deviations in temperature, which allows us to investigate what other parameters may have driven the anomalies.

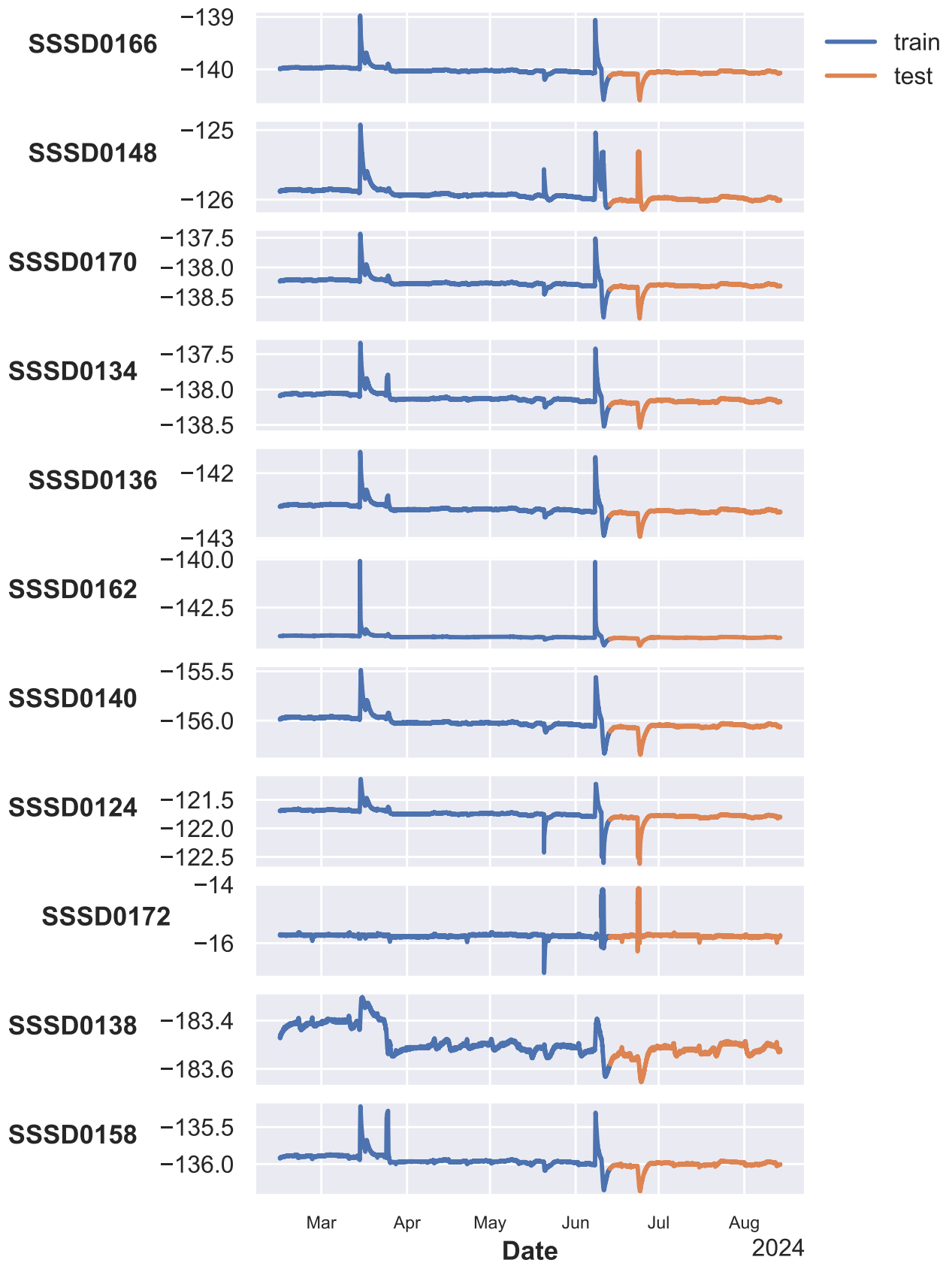

Fig. 2. Plots of the 11 temperature parameters over the studied period indicating which data were used for training and testing, respectively

We identified the top three anomalies for each parameter as the 10-day spans with the highest mean value. This choice was made based on initial inspection of the data and will be automated in later iterations. However, for this first investigation with a limited number of parameters that are all temperatures, it proved sufficient.

### *2.3 SHAP Analysis for Anomaly Interpretation*

After identifying the top anomalies for each temperature parameter, we apply SHAP (Shapley Additive Explanations) to interpret the predictions made by Model 2. SHAP values provide a way to attribute the contributions of the covariates to the model's predictions, helping to identify which covariates had the greatest impact on the temperature residuals and therefore on the temperature anomalies.

For each identified anomaly, we compute SHAP values over a 10-day window around the anomaly. This analysis allows us to interpret which covariates were most influential for the residual and thus anomaly prediction during the anomalous period. By understanding the contribution of each covariate to the residuals, we can gain insight into the potential causes of the temperature anomalies.

This two-model approach, combining temperature prediction with residual-based anomaly prediction, allows us have separate models for understanding the parameters behaviour but also the underlying relationship with other parameters and anomalies.

## 3. Results

In this section, we present the results of our anomaly detection and interpretation methodology applied to the temperature telemetry data collected from the Euclid spacecraft. Given space limitations we focus on two temperature parameters, SSSD0166 (baseplate temperature near VIS instrument) and SSSD0172 (VIS instrument radiator temperature), and provide a detailed analysis of the top anomalies detected for them. For each parameter, we present the anomaly detection results, including SHAP-based interpretability to understand the role of various covariates in predicting the temperature residuals and anomalies.

### *3.1 SSSD0166: Anomaly Detection and SHAP Analysis*

For the temperature parameter SSSD0166, we identified the most significant anomaly occurring between June 17 and June 27, 2024. This event is related to a VIS instrument safe mode activation where thermal dissipation suddenly dropped until operations were recovered following a reboot. Figure 6 shows the predicted temperature values compared to the actual temperature values, highlighting the top three anomalies detected in the second half of June. Additionally, in the bottom plot anomaly and predicted anomaly score are shown, which clearly stand out during the detected anomaly.

In order to interpret the causes of these anomalies, we applied SHAP analysis to the residual predictions from Model 2, which was trained on the covariate parameters. Figure 7 provides a SHAP summary plot over the entire time span for SSSD0166, indicating the relative importance of the covariates in driving the residual predictions.

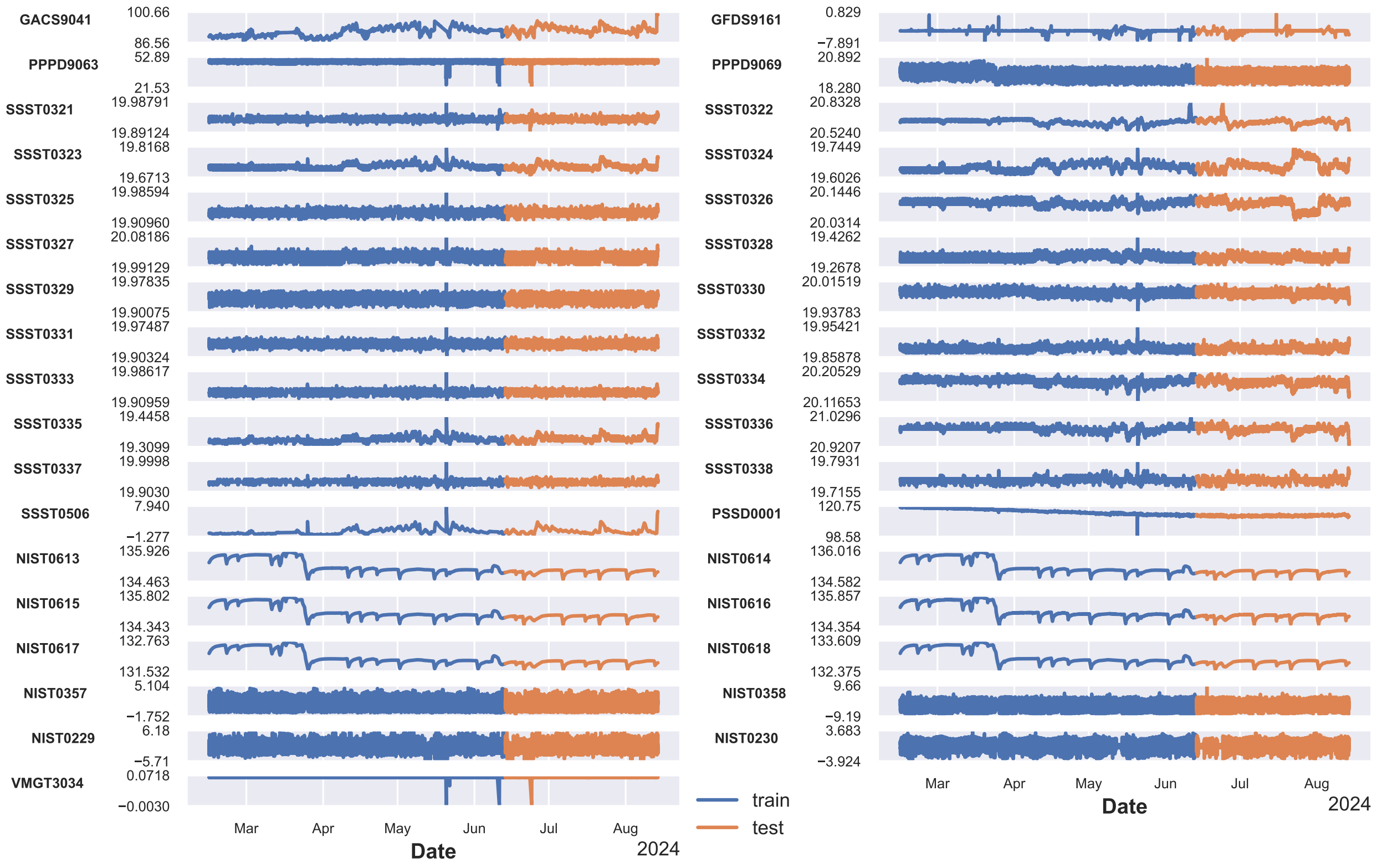

Fig. 3. Plots of the 35 covariate parameters over the studied period indicating which data were used for training and testing, respectively

The SHAP analysis revealed that the most important covariates for predicting temperature anomalies in SSSD0166 were the current temperature values of NIST0617, as well as SSST0322 and SSST0506 values from a day prior. Interestingly, the VIS instrument shutter motor currents (VMGT3034) had the largest individual SHAP values but not on average. For all parameters, larger values were associated with more anomalous predictions.

Focusing on the top anomaly from June 17 to June 27, we provide a detailed SHAP heatmap in Figure 8. This heatmap highlights the contributions of each covariate during this anomalous period. The most significant contributors to the anomaly prediction were SSST0322 one day prior and VMGT3034 10 and 30 minutes before. We further analyse this anomaly by visualising these covariates and their respective SHAP values against the behaviour of SSSD0166 and anomaly predictions over this interval in Figure 9.

The SHAP plots clearly show the relationship between the temperature parameter and the covariates, with the VMGT3034 and SSST0322 parameters showing significant changes during the anomalous period. Note that SHAP values for SSST0322 lag behind given the one day delay likely caused by time delay due to heat flow.

### *3.2 SSSD0172: Anomaly Detection and SHAP Analysis*

For SSSD0172, we similarly identified an important anomaly occurring between June 15 and June 25, 2024. Figure 10 shows the predicted temperature values compared to the actual values, with the top three anomalies highlighted and predicted and computed k-means anomaly values in the bottom.

SHAP analysis revealed that VMGT3034, GFDS9161 (alpha angle), and SSST0506 from a day prior were the most important covariates for predicting temperature anomalies in SSSD0172. Figure 11 shows the SHAP sum-mary plot for SSSD0172, highlighting the relative importance of these covariates. Notably NIST06xx play little role in the anomalies in SSSD0172 in line with the lack of correlation observed in Figure 5.

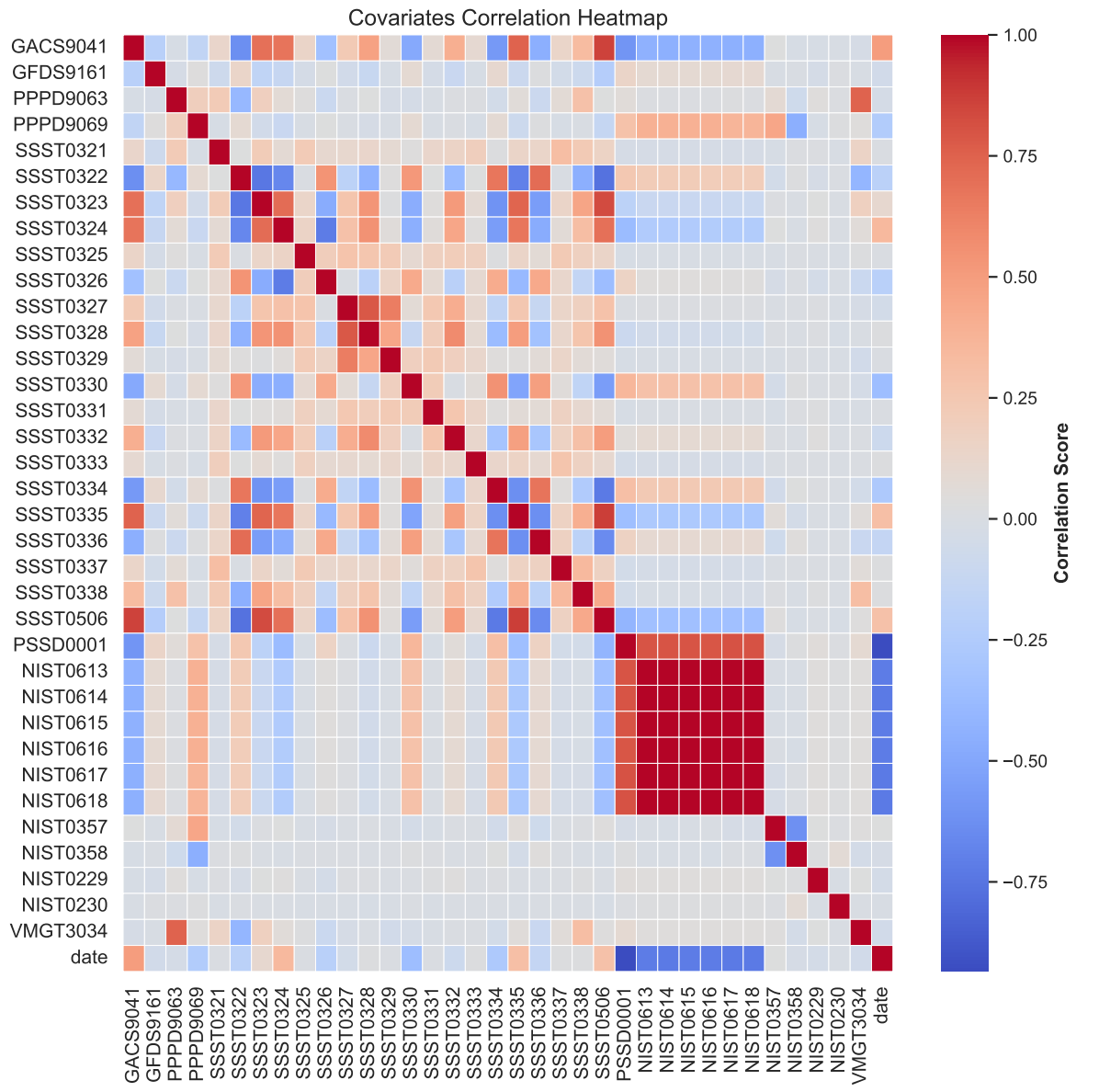

Fig. 4. Covariate correlation heatmap

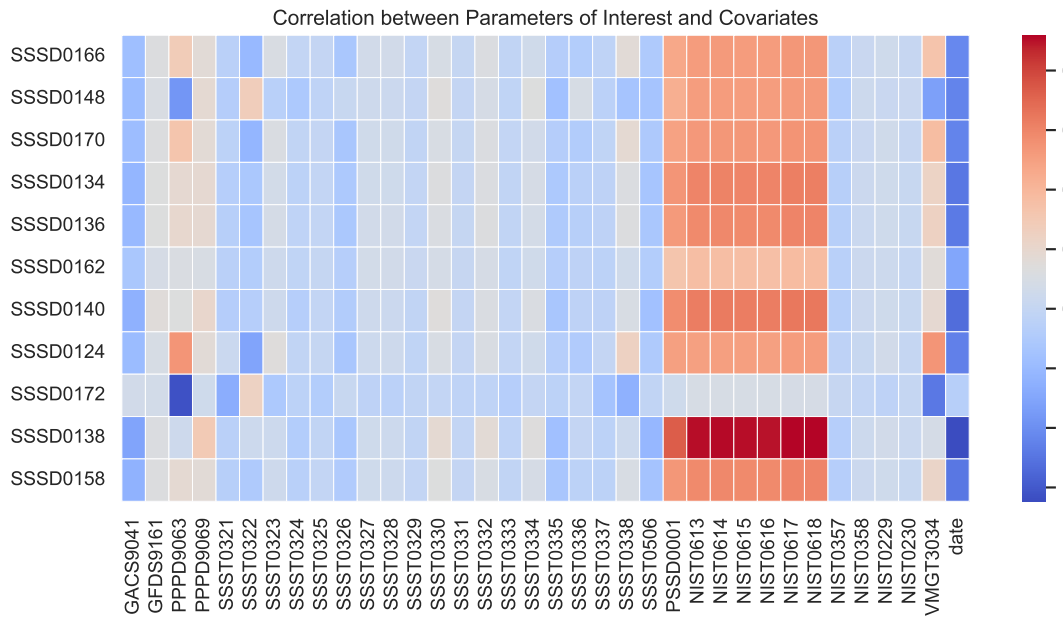

Fig. 5. Observed temperature parameters' correlation with covariates as heatmap

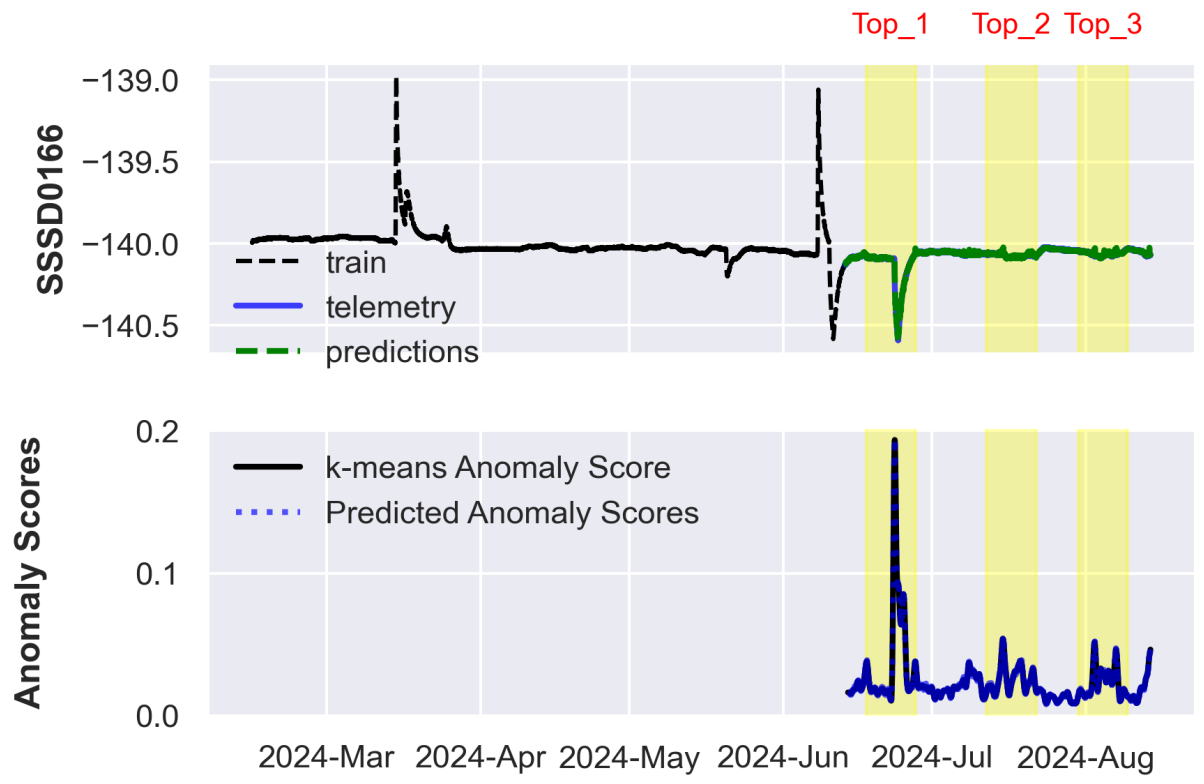

Fig. 6. Temperature predictions for SSSD0166 with the top 3 anomalies highlighted. The largest anomaly occurred between June 17 and June 27.

A detailed SHAP heatmap of the June 15-25 anomaly is shown in Figure 12, which highlights the key contributors to the anomaly prediction during this interval. VMGT3034 and the power values of the VIS CPDU unit (PPPD9063) were particularly influential, along with the alpha angle (GFDS9161) from a day before.

Figure 13 provides a closer look at the top three covariates (VMGT3034, PPPD9063, and GFDS9161) and their SHAP values over the June 15-25 period, showing a clear relationship between these parameters and the temperature anomalies observed in SSSD0172. Note that given the 24 hour delay GFDS9161 SHAP values spike likely at jump a day before the anomaly and small slope changes in GFDS9161 before the drop at the end of the anomaly seem to already influence the model via mildly negative SHAP values.

Overall, these results clearly show the successful identification of significant relationships in the parameters.

## 4. Discussion

In this work we demonstrate the successful application of machine learning techniques to detect and interpret temperature anomalies in the Euclid spacecraft's telemetry data. Through the use of two XGBoost models building on each other, we were able to not only identify temperature anomalies but also study their possible causes by identifying key covariates influencing the temperature deviations. The results from our analysis provide valuable insights and suggest further opportunities for enhancing this methodology to different parameters and in future space missions.

### *4.1 Interpretation of Results*

The combined model approach proved highly effective at detecting significant temperature anomalies. By separating the temperature prediction from the anomaly inference, we could identify not only when an anomaly occurred but also gain insights into which covariates may have driven these deviations. A normal predictive model would not have been able to do so as its SHAP values would identify cause for the prediction not for anomalies. The results highlighted the influence of certain covariates, such as the VIS instrument shutter motor current (VMGT3034), solar angles (GACS9041 and GFDS9161), and other connecting ring or service module temperatures (SSST0322 and SSST0506). These covariates often displayed a strong influence on temperature behaviour in the studied parameters, suggesting potential physical relationships between these subsystems. It is important to stress, we aim to reveal relationships present in the data, but the detailed physical interpretation from root cause to consequences in system response is beyond the scope of this initial study. Sensitivity of the PLM baseplate thermal response to attitude variations or to VIS internal heat load changes are known effects, but these relationships were not yet explored with the models presented.

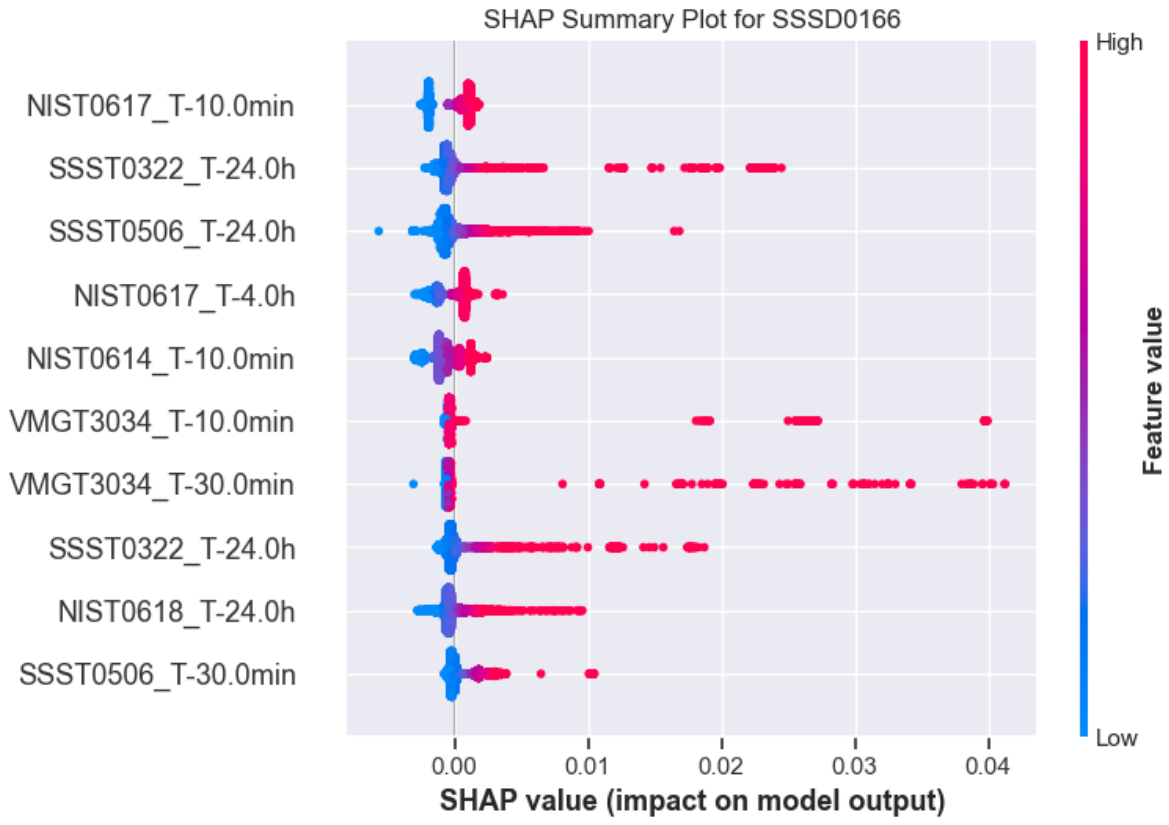

Fig. 7. SHAP summary plot for SSSD0166, showing the most important covariates for predicting temperature anomalies.

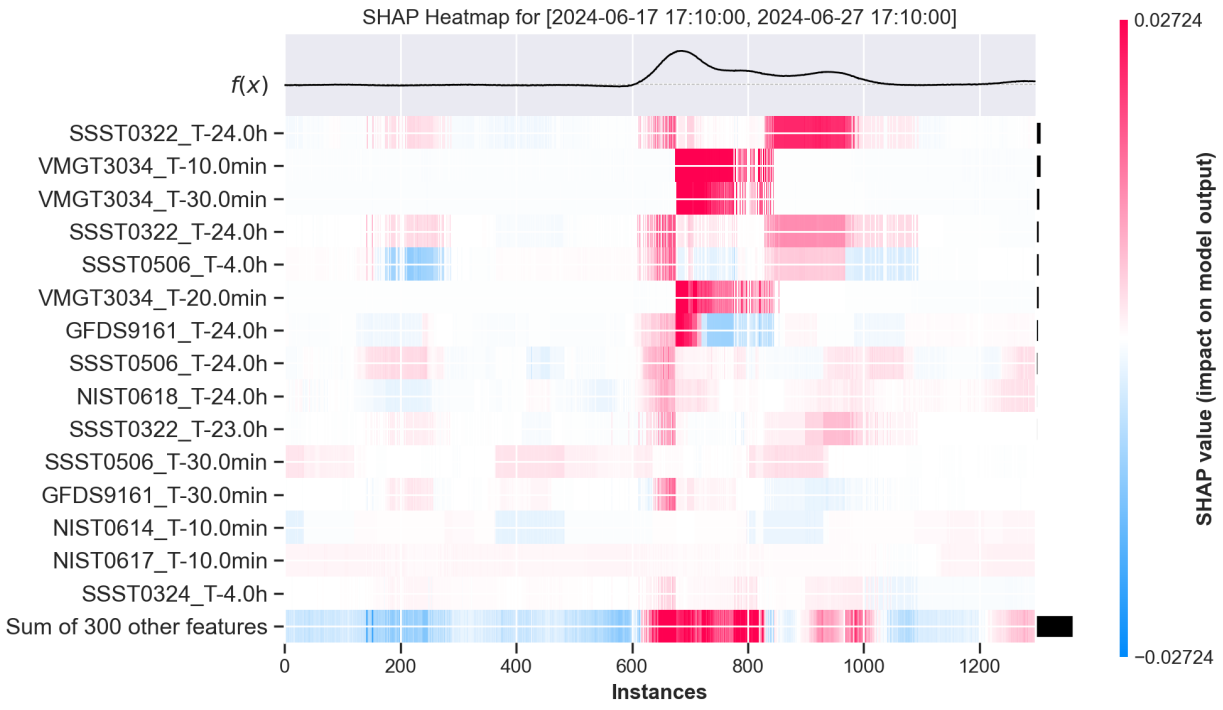

Fig. 8. SHAP heatmap for SSSD0166 during the June 17-27 anomaly window. SSST0322 and VMGT3034 were the most influential covariates.

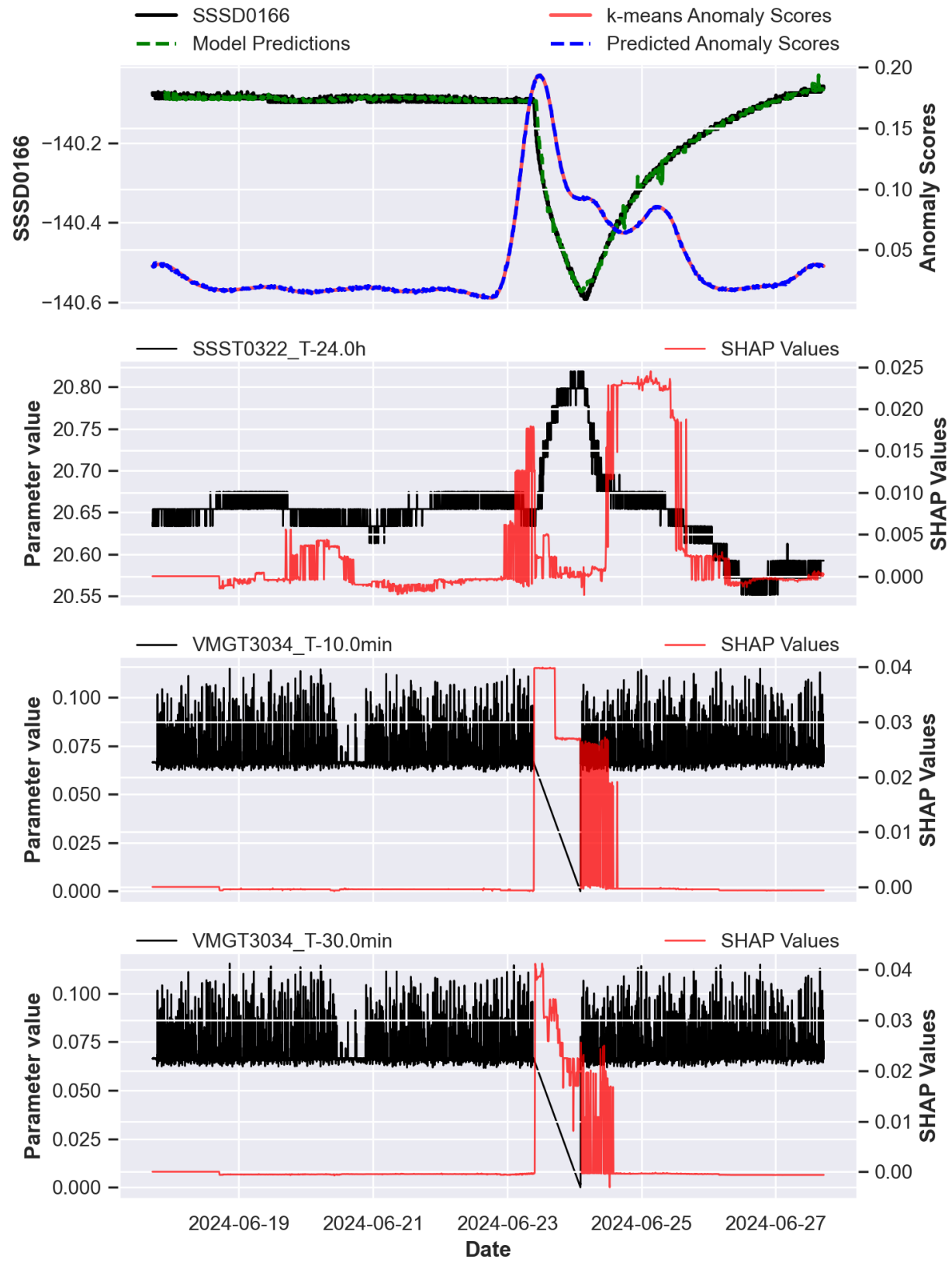

Fig. 9. Detailed SHAP analysis for SSSD0166 between June 17 and June 27. The top three covariates (SSST0322, VMGT3034, and SSST0506) and their SHAP values are shown.

However, it is important to note that while the XGBoost model effectively selected key features for predicting anomalies, this does not imply that other covariates could not also be good predictors and hence physically relevant. The model identifies strong relationships for the dataset used, but additional features or other time lags may also have predictive power. Therefore, future work should conduct more detailed analyses using ablation studies, ensemble models, and other techniques to robustly identify all potential relationships between parameters and anomalies. These more rigorous approaches will help uncover hidden covariate dependencies that were not immediately evident in the current study.

### *4.2 Computational Efficiency*

A notable advantage of this approach is its computational efficiency. Each temperature parameter could be processed within approximately two minutes, making the method suitable for near-real-time anomaly detection. This level of performance is particularly promising to quickly adapt operations to avert a negative impact on scientific return of the mission. Thus, these models can help rapidly detecting and studying anomalies to ensure

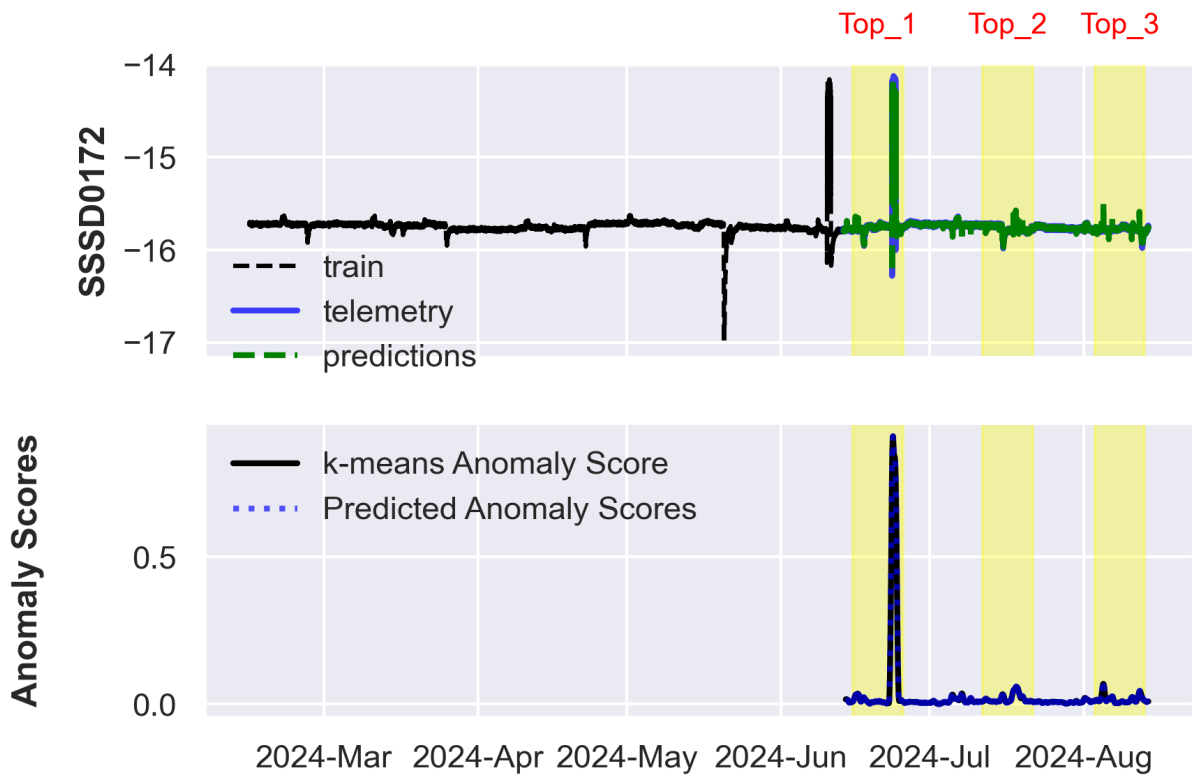

Fig. 10. Temperature predictions for SSSD0172 with the top 3 anomalies highlighted. The largest anomaly occurred between June 15 and June 25.

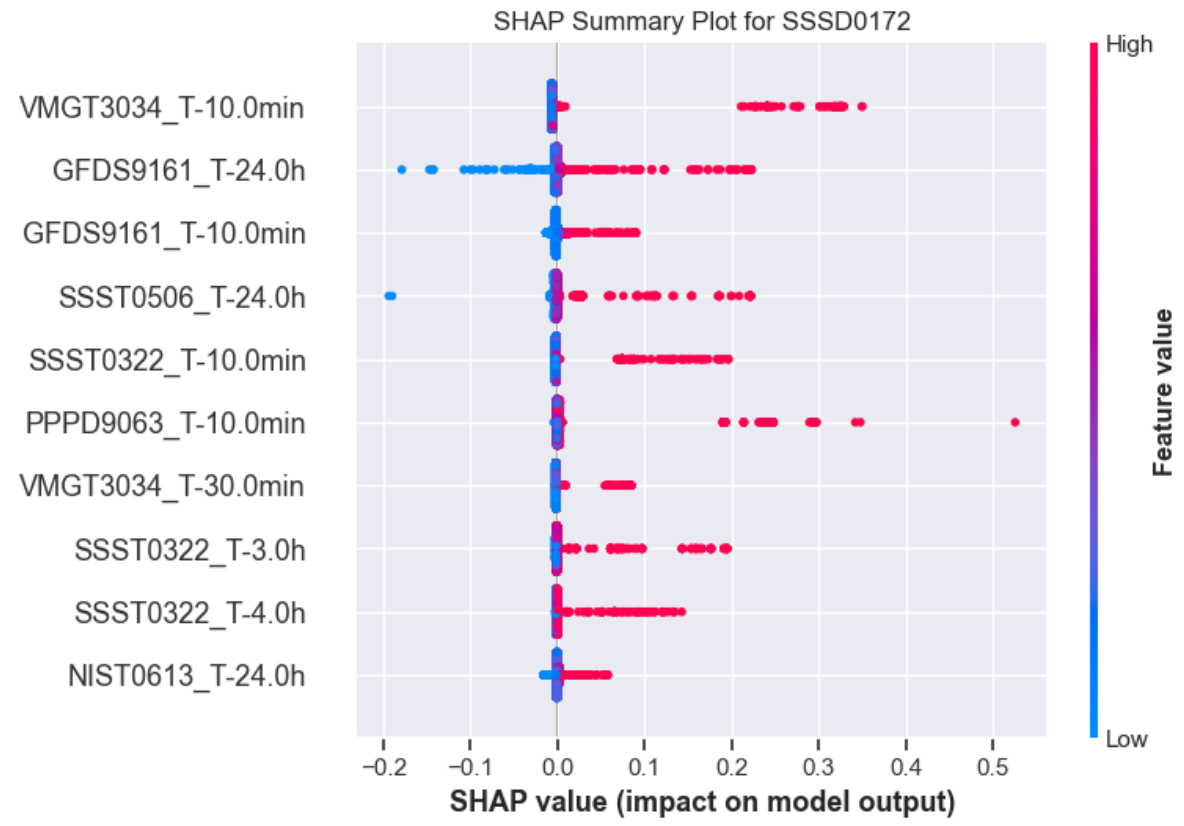

Fig. 11. SHAP summary plot for SSSD0172, showing the most important covariates for predicting temperature anomalies.

that mission operators can respond quickly, thus improving overall mission safety and performance.

### *4.3 Insights from SHAP Analysis*

The SHAP analysis provided crucial insights into the relationships between temperature anomalies and the covariate parameters. By visualising the SHAP values, we can identify not only the most important covariates but also how their influence changed over time. For instance, parameters like VMGT3034 showed strong short-term influences during anomalies, while others, such as SSST0322, exhibited lagged effects, likely due to the time it takes for heat to propagate through the spacecraft's structure. These patterns offer valuable insights into the underlying physical processes driving the anomalies.

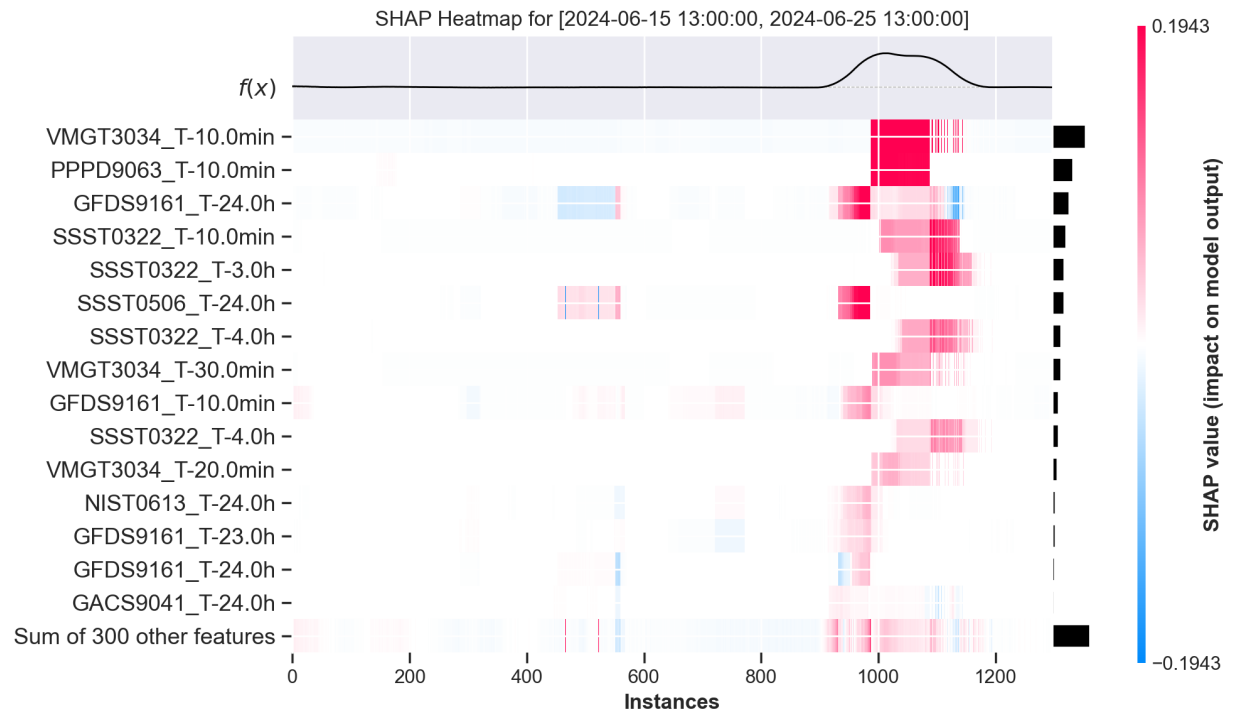

Fig. 12. SHAP heatmap for SSSD0172 during the June 15-25 anomaly window. VMGT3034 and PPPD9063 were the most important covariates.

However, one limitation of SHAP analysis is that it does not directly infer causality. While SHAP values allow us to attribute changes in the temperature residuals to specific covariates, they do not necessarily explain why these covariates are influential. Future work should explore more rigorous approaches to causal inference, such as counterfactual explanations or other interpretable machine learning techniques, to complement the insights gained from SHAP.

### *4.4 Potential Future Work*

To build upon the work presented in this study, several avenues for future research and development are conceivable. First, it would be useful to apply this approach to known anomalies in the telemetry data to validate its feasibility in a more systematic and rigorous fashion that was beyond the scope of this initial investigation. By testing the model's ability to predict known issues, we could better assess its reliability and potential for future applications. Secondly, we aim to perform more detailed analyses using ablation studies and ensemble models to systematically identify all covariates that influence temperature anomalies, rather than relying solely on individual XGBoost models. Additionally, it would be interesting to explore the application of this methodology to other types of telemetry data, such as voltages, currents or spacecraft health-related measures to evaluate its generalisability across different spacecraft systems. One core component missing at the moment is the integration of operational context (e.g., spacecraft events, mission timeline) to better contextualise anomalies and improve the interpretability of the results. For instance, linking anomalies to specific spacecraft events, such as science instrument activities could enhance the understanding of the driving factors behind the anomalies.

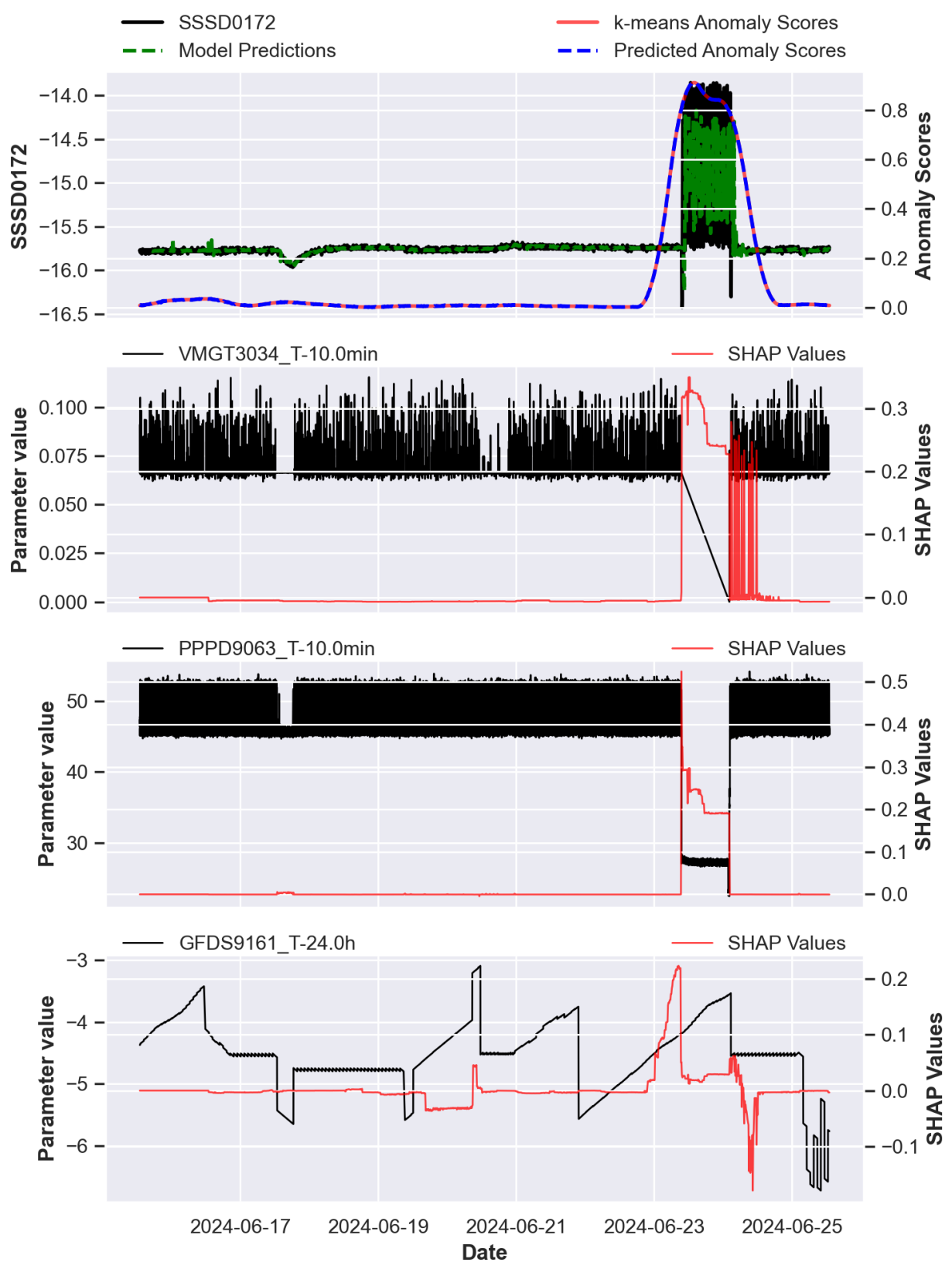

Fig. 13. Detailed SHAP analysis for SSSD0172 between June 15 and June 25. The top three covariates (VMGT3034, PPPD9063, and GFDS9161) and their SHAP values are shown.

### *4.5 Implications for Future Missions*

The combination of predictive anomaly detection and SHAP-based interpretation demonstrated in this study represents a scalable and adaptable approach for handling large-scale telemetry data in modern spacecraft. Given the increasing data volumes and complexity of space missions, machine learning methods like the one presented here offer a viable path to automate telemetry monitoring, detect anomalies early, and provide mission operators with actionable insights into potential causes.

This method shows great promise for Euclid in particular, where thermal stability is crucial to the success of the mission's scientific objectives. However, the approach can also be applied to other spacecraft, both within and beyond ESA, to monitor a variety of telemetry parameters and subsystems. By adopting machine learning-based monitoring and interpretability methods, future missions can enhance their operational resilience, improve safety, and ultimately increase mission efficiency.

## 5. Conclusions

This work has demonstrated the effective use of machine learning to detect and interpret temperature anomalies in Euclid's telemetry data. By combining two predictive models—one focused on temperature prediction and the other on anomaly interpretation based on covariates—we were able to identify key deviations and provide insights into their underlying causes using SHAP (Shapley Additive Explanations).

Key findings include the identification of important covariates influencing temperature anomalies. The methodology proved efficient, processing each parameter in approximately two minutes, suitable for near-real-time applications.

While this approach has shown promise for Euclid's temperature monitoring, future work will explore additional relationships through ablation studies and ensemble models. Applying this methodology to other spacecraft subsystems and known anomalies could further validate its effectiveness.

Overall, we demonstrate that the use of machine learning and explainable AI in spacecraft telemetry offers significant potential for improving mission operations, providing valuable automated insights, and enhancing overall mission efficiency.

## Acknowledgements

The authors would like to thank Tristan Dijkstra, Stefano Speretta and the XMM-Newton team for valuable discussions around machine learning for telemetry anomalies.